\documentclass[pdflatex,sn-mathphys-num]{sn-jnl}% Math and Physical Sciences Numbered Reference Style
\usepackage{graphicx}%
\usepackage{multirow}%
\usepackage{amsmath,amssymb,amsfonts}%
\usepackage{amsthm}%
\usepackage{mathrsfs}%
\usepackage[title]{appendix}%
\usepackage{xcolor}%
\usepackage{textcomp}%
\usepackage{manyfoot}%
\usepackage{booktabs}%
\usepackage{algorithm}%
\usepackage{algorithmicx}%
\usepackage{algpseudocode}%
\usepackage{listings}%
\usepackage{array}

\theoremstyle{thmstyleone}%
\theoremstyle{thmstyletwo}%

\theoremstyle{thmstylethree}%

\begin{document}

\title[Article Title]{CT-Bench: A Benchmark for Multimodal Lesion Understanding in Computed Tomography}

\author[1]{\fnm{Qingqing} \sur{Zhu}}
\equalcont{These authors contributed equally to this work.}
\author[1]{\fnm{Qiao} \sur{Jin}}
\equalcont{These authors contributed equally to this work.}

\author[1]{\fnm{Tejas S.} \sur{Mathai}}
\author[1]{\fnm{Yin} \sur{Fang}}
\author[1]{\fnm{Zhizheng} \sur{Wang}}
\author[1]{\fnm{Yifan} \sur{Yang}}
\author[1,2]{\fnm{Maame} \sur{Sarfo-Gyamfi}}
\author[1]{\fnm{Benjamin} \sur{Hou}}
\author[1]{\fnm{Ran} \sur{Gu}}
\author[1]{\fnm{Praveen T. S.} \sur{Balamuralikrishna}}
\author[3]{\fnm{Kenneth C.} \sur{Wang}}
\author[1]{\fnm{Ronald M.} \sur{Summers}}
% \author[1]{\fnm{Zhiyong} \sur{Lu}$^{*}$}\email{zhiyong.lu@nih.gov}

% \footnotetext[1]{* Corresponding author}

% \author[1]{\fnm{Zhiyong} \sur{Lu}}
% \email{Corresponding author: zhiyong.lu@nih.gov}

\author[1]{\fnm{Zhiyong} \sur{Lu}$^{*}$}
\footnotetext[1]{* Corresponding author: zhiyong.lu@nih.gov}
% \footnotetext{*Corresponding author: zhiyong.lu@nih.gov}

% \author[1]{\fnm{Zhiyong} \sur{Lu}\corref{cor1}}\email{zhiyong.lu@nih.gov}

% \cortext[cor1]{Corresponding author}

% \author*[1,2]{\fnm{First} 
% \sur{Author}}\email{iauthor@gmail.com}



% \affil*[1]{\orgdiv{Department}, \orgname{Organization}, \orgaddress{\street{Street}, \city{City}, \postcode{100190}, \state{State}, \country{Country}}}

% \affil[2]{\orgdiv{Department}, \orgname{Organization}, \orgaddress{\street{Street}, \city{City}, \postcode{10587}, \state{State}, \country{Country}}}

% \affil[3]{\orgdiv{Department}, \orgname{Organization}, \orgaddress{\street{Street}, \city{City}, \postcode{610101}, \state{State}, \country{Country}}}


\abstract{Artificial intelligence (AI) can automatically delineate lesions on computed tomography (CT) and generate radiology report content, yet progress is limited by the scarcity of publicly available CT datasets with lesion-level annotations.  To bridge this gap, we introduce CT-Bench, a first-of-its-kind benchmark dataset comprising two  components:
a Lesion Image \& Metadata Set containing 20,335 lesions from 7,795 CT studies with bounding boxes, descriptions, and size information, and a multitask visual question answering benchmark with 2,850 QA pairs covering lesion localization, description, size estimation, and attribute categorization. Hard negative examples are included to reflect real-world diagnostic challenges.
We evaluate multiple state-of-the-art multimodal models—including vision-language and medical CLIP variants—by comparing their performance to radiologist assessments, demonstrating the value of CT-Bench as a comprehensive benchmark for lesion analysis. Moreover, fine-tuning models on the Lesion Image \& Metadata Set yields significant performance gains across both components, underscoring the clinical utility of CT-Bench.
}




\maketitle

\section{Introduction}\label{sec1}

% \section{Introduction}

Medical imaging plays a pivotal role in modern healthcare diagnostics and treatment, with computed tomography (CT) standing out for its wide application across multiple regions of the body, such as the chest, abdomen, and skeletal system. Advancements in artificial intelligence (AI) have revolutionized medical image analysis, diagnosis, and treatment planning \cite{tian2024opportunities,hou2025one}. These innovations rely on large, annotated radiological datasets to train robust AI models \cite{sun2023scoping, zhu2025well}. However, a major challenge in AI-driven CT analysis is the shortage of large-scale, high-quality annotated datasets, primarily due to the immense labor and domain expertise required to produce reliable clinical annotations \cite{zhu2024decomposing}.

While a number of datasets have supported AI research in radiology, most exhibit important limitations for CT-based multimodal learning. For example, DeepLesion \cite{yan2017deeplesion} provides over 32,000 lesion annotations but does not include textual descriptions. CT-RATE \cite{hamamci2024foundation} supports multimodal learning with 3D chest CTs and full radiology reports, yet lacks 2D slice-level, lesion-specific annotations. Other resources such as ROCOv2 \cite{ruckert2024rocov2}, PMC-OA \cite{lin2023pmc}, and MedICaT \cite{subramanian2020medicat} rely on captions extracted from biomedical literature, which often lack clinical specificity. Meanwhile, high-quality multimodal datasets have been developed for chest X-ray, such as MIMIC-CXR \cite{johnson2019mimic}, but no comparable large-scale multimodal dataset currently exists for CT. Table~\ref{tab:medical_datasets} summarizes representative medical imaging datasets used for multimodal learning.

\begin{figure*}[!hbt]
    \centering
\includegraphics[width=\linewidth]{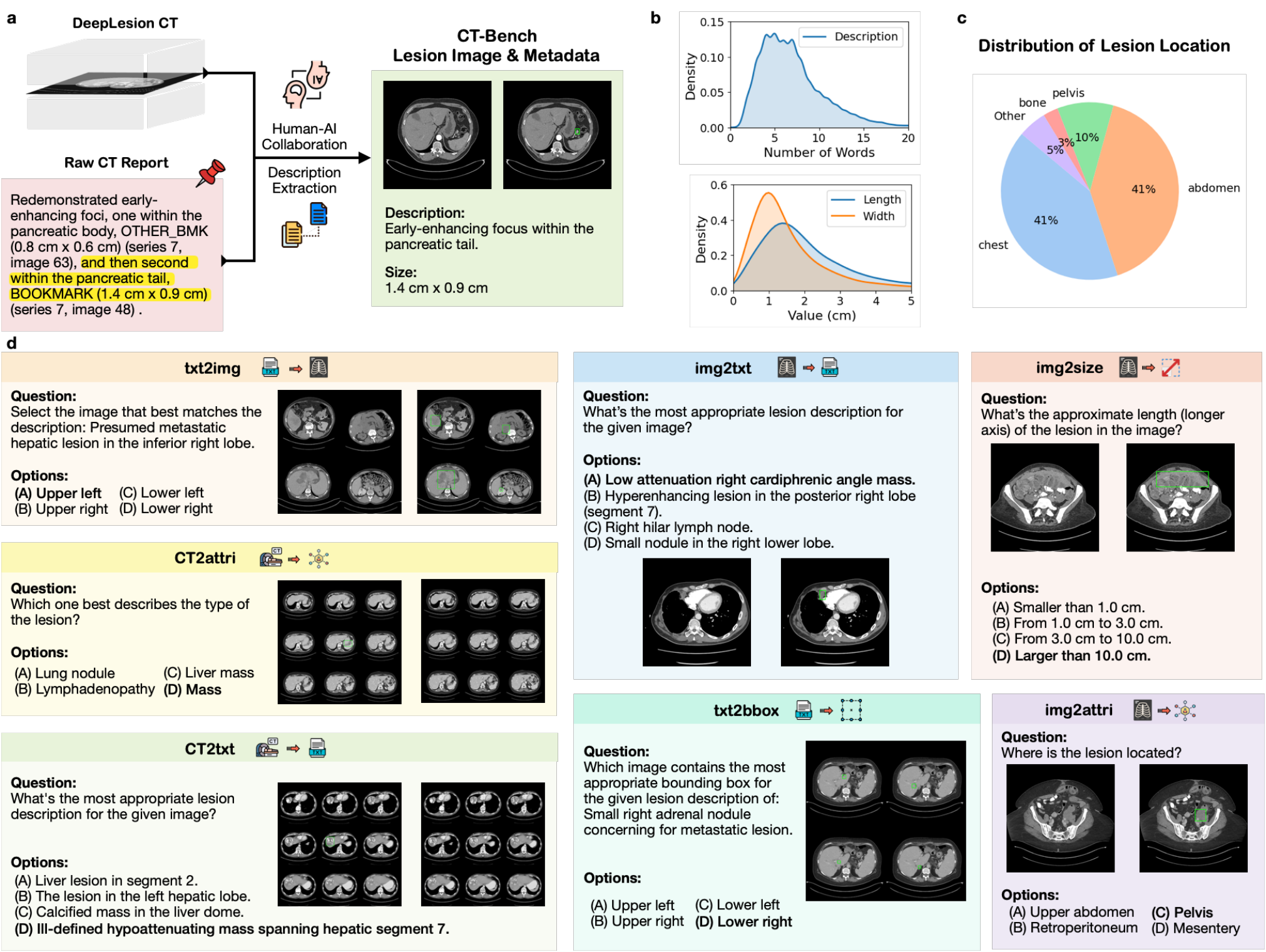}

    \caption{Overview of CT-Bench. (a) Example of data annotation, showing the transformation of original reports from PACS into detailed descriptions enriched with size information for CT-Bench: Lesion Image \& Metadata Set. (b) Analysis of description: word count and value distribution. (c) Distribution of lesion locations. (d) CT-Bench: components of the QA benchmark cases.}
    \label{fig:template}
\end{figure*}

In contrast, we introduce \textbf{CT-Bench}, a first-of-its-kind resource designed to support multimodal AI for clinical CT interpretation. The first component, \textbf{CT-Bench: Lesion Image \& Metadata Set}, pairs 2D CT slices and optional lesion-centered 3D sub-volumes with detailed textual annotations extracted directly from hospital Picture Archiving and Communication Systems (PACS). Through the curation process illustrated in Figure~\ref{fig:template}(a), raw clinical reports are transformed into structured, high-quality data, including lesion descriptions, size measurements, and bounding box (BBox) annotations, making them suitable for multimodal AI development.

Beyond dataset construction, CT-Bench also provides a comprehensive benchmark for evaluating multimodal reasoning in CT. Building on the Lesion Image \& Metadata Set, we further introduce the \textbf{CT-Bench: QA Benchmark Component}, a multitask visual question answering (VQA) benchmark designed to evaluate and advance multimodal AI in CT-based clinical applications. Several medical VQA datasets have been proposed, including VQA-RAD \cite{lau2018dataset}, PathVQA \cite{he2020pathvqa}, SLAKE \cite{liu2021slake}, VQA-Med \cite{ben2021overview}, and PMC-VQA \cite{zhang2023pmc}. However, these datasets are often limited in size, modality coverage, or annotation depth, and many focus primarily on 2D images or rely on open-ended answer formats that complicate standardized evaluation. Table~\ref{tab:qa_datasets} summarizes representative medical VQA benchmarks.

Most existing VQA datasets do not support multi-slice CT reasoning and rarely include lesion-specific bounding boxes, lesion size information, or hard negative examples. CT-Bench: QA Benchmark addresses these limitations with 2.8k multiple-choice QA pairs spanning seven lesion-level tasks on multi-slice CT images, incorporating bounding boxes and radiologist-validated hard negatives to enable rigorous and clinically grounded evaluation. Specifically, CT-Bench supports seven core tasks: single-slice lesion captioning, text-guided slice retrieval, lesion localization with bounding boxes, lesion size estimation, single-slice attribute classification, multi-slice lesion captioning, and multi-slice attribute classification.

Our key contributions are summarized as follows:
(1) \textbf{CT-Bench: Lesion Image, 3D Sub-volume \& Metadata Set}, a high-quality multimodal CT dataset comprising 20,335 lesions across 7,795 studies with PACS-derived annotations and lesion-centered 3D sub-volumes.
(2) \textbf{CT-Bench: QA Benchmark Component}, a novel VQA benchmark supporting seven lesion analysis tasks and integrating hard negative examples for rigorous evaluation.
(3) A comprehensive evaluation of state-of-the-art multimodal AI models, including general vision-language models, medical vision-language models, and medical CLIP models.
(4) Demonstration that fine-tuning existing models on the Lesion Image \& Metadata Set leads to substantial performance gains across both components of CT-Bench, highlighting its importance for building multimodal AI systems.

\section{Results}\label{sec2}

\subsection{Image Caption Generation on the Lesion Image \& Metadata Set} 

Table \ref{performance_comparison} provides a comparative analysis of various models' performance. We find that: (1) Models generally perform better with BBox (w/) than without BBox (w/o) for most metrics. 
(2) Among all non-tuned models, \textbf{Dragonfly} and \textbf{Gemini} demonstrated best performance. However,  the performances are still low, indicating significant challenges in integrating multimodal information. (3) \textbf{Fine-Tuned RadFM(w)} achieves the highest scores across all metrics. These results significantly outperform all baseline models and validate the effectiveness of model adaptation using our dataset.
The \textbf{Fine-Tuned RadFM(w/o)} model also performs competitively. Although its performance is slightly lower than that of its BBox-enhanced configurations, this result demonstrates that fine-tuning alone significantly boosts performance—even without spatial localization inputs.  CLIP-style models are excluded from this comparison, as they are not capable of performing the image caption generation task.

\subsection{Case Study: Image Captioning Results on CT-Bench}

Table~\ref{tab:model_comparison} presents a qualitative case study comparing image captioning outputs on the CT-Bench: Lesion Image \& Metadata Set, illustrating the challenges of generating accurate and clinically meaningful captions for medical images. Unlike general image captioning, lesion description in CT requires precise localization, concise yet specific descriptions, and avoidance of irrelevant or incorrect findings.

Baseline models exhibit several common failure modes. For example, \textbf{LLaVA-Med} and \textbf{Gemini} misidentify the lesion location, incorrectly describing it as being in the right lung rather than the left lower lobe. \textbf{RadFM} introduces extraneous and incorrect details, such as referencing multiple nodules and a draining vessel that are not present in the image.

In contrast, the fine-tuned \textbf{RadFM} (w/ BBox) model produces more accurate and clinically relevant captions, correctly identifying lesion location and avoiding hallucinated findings. These examples qualitatively demonstrate the benefit of fine-tuning with structured annotations.

% \begin{table*}[t]\small
% \centering
% {
% \begin{tabular}{lcccccccccccccccc}
% \toprule
% \textbf{Model} & 
% \multicolumn{2}{c}{\textbf{BERT P}} & 
% \multicolumn{2}{c}{\textbf{BERT R}} & 
% \multicolumn{2}{c}{\textbf{BERT F1}} & 
% \multicolumn{2}{c}{\textbf{ClinicalBERT CosSim}} \\
% \cmidrule(lr){2-3} \cmidrule(lr){4-5} \cmidrule(lr){6-7} \cmidrule(lr){8-9}
% & \textbf{w/o} & \textbf{w/} 
% & \textbf{w/o} & \textbf{w/} 
% & \textbf{w/o} & \textbf{w/} 
% & \textbf{w/o} & \textbf{w/} \\
% \midrule
% RadFM & 0.080 & 0.082 & 0.166 & 0.171 & \textbf{0.122} & 0.126 & \textbf{0.838} & 0.840 \\
% Dragonfly & 0.012 & \textbf{0.211} & 0.112 & 0.228 & 0.061 & \textbf{0.219} & 0.817 & \textbf{0.854} \\
% Llava\_med & \textbf{0.027} & 0.04 & 0.152 & 0.165 & 0.089 & 0.102 & 0.813 & 0.812 \\
% GPT-4V & -0.089 & 0.048 & 0.085 & 0.236 & -0.003 & 0.14 & 0.830 & 0.825 \\
% Gemini & 0.016 & 0.067 & \textbf{0.209} & \textbf{0.257} & 0.111 & 0.160 & 0.834 & 0.826 \\
% \hline
% TunedRadFM(w/o) & \underline{0.464} & 0.464 & 0.294 & 0.293 & 0.377 & 0.376 & \underline{0.916} & 0.916\\
% TunedRadFM(w) & 0.447 & \underline{0.466} & \underline{0.313} & \underline{0.319} & \underline{0.379} & \underline{0.386} & 0.907 & \underline{0.919} \\

% \bottomrule
% \end{tabular}
% }
% \caption{Summary of performance for models with and without BBox. Abbreviations: \textbf{w/o} = Without BBox, \textbf{w/} = With BBox.}
% \vspace{-0.05in}
% \label{performance_comparison}
% \end{table*}

\subsection{Results on CT-Bench  QA Benchmark for all models} 
The CT-Bench dataset serves as both a challenging benchmark and a comprehensive framework for evaluating the strengths and weaknesses of AI models across a variety of tasks and configurations.  The evaluation results are in Table \ref{tab:model_performance_weighted} (validation results are provided in the Appendix)
From this table we can find that:  (1) \textbf{BiomedCLIP} itself consistently led among untuned models, with an average score of \textbf{41.00\%} (w/ BBox) and \textbf{34.89\%} (w/o BBox), securing top performance in key language-grounding tasks like \texttt{Img2txt} and \texttt{Context2txt}. This shows robust visual-language alignment even without spatial cues.
(2) \textbf{Gemini} and \textbf{Dragonfly} showed promising performance in visual-spatial tasks. Gemini excelled in \texttt{Txt2bbox} with BBox (36.00\%) and \texttt{Img2size} (50.00\%), suggesting strong capabilities in grounding textual queries to visual regions. Dragonfly, while not dominating any single task, demonstrated consistently balanced performance across tasks, particularly in attribute-related questions (e.g., 49.00\% in \texttt{Img2attrib} w/o BBox).
(3) \textbf{GPT-4V} delivered moderate performance across most tasks, with noticeable strength in attribute recognition (44.00\% in \texttt{Img2attrib} w/ BBox), but struggled with tasks requiring spatial grounding, especially \texttt{Txt2bbox}. \textbf{LLaVA-Med} and \textbf{RadFM}, both open-source vision-language models adapted for the medical domain, underperformed significantly, often scoring near or below random chance—suggesting a lack of alignment with the complex multimodal nature of CT-based clinical tasks.
(4) A striking outcome was that RadFM, after fine-tuning solely on the image caption dataset, scored zero on all CT-Bench QA tasks. This collapse suggests catastrophic forgetting \cite{goodfellow2013empirical} — a common pitfall in neural networks where new learning overwrites prior task knowledge. Unlike humans, models fine-tuned narrowly can lose all prior task competence.
 (5) \textbf{Fine-tuned BiomedCLIP} emerged as the strongest performer overall, achieving the highest average accuracy of \textbf{62.00\%} BBox when utilizing Binputs, substantially outperforming all other models. It also retained strong performance without BBox inputs (\textbf{41.44\%}), surpassing the base BiomedCLIP and all other non-tuned models. 
 These results underscore the significance of our CT-Bench  dataset in facilitating the development of improved models. 
  
 % \qiao{I think after texttt these subtasks generally look better. Maybe you can do so for all occurrences of img2txt, Context2txt, etc...}
% \documentclass{article}
% \usepackage{booktabs}
% \usepackage{xcolor} % For text color

% \begin{document}
\subsection{Effect of Bounding Box and Task Difficulty}
\begin{figure}

    \vspace{-10pt}  
    \centering
    \includegraphics[width=0.48\textwidth]{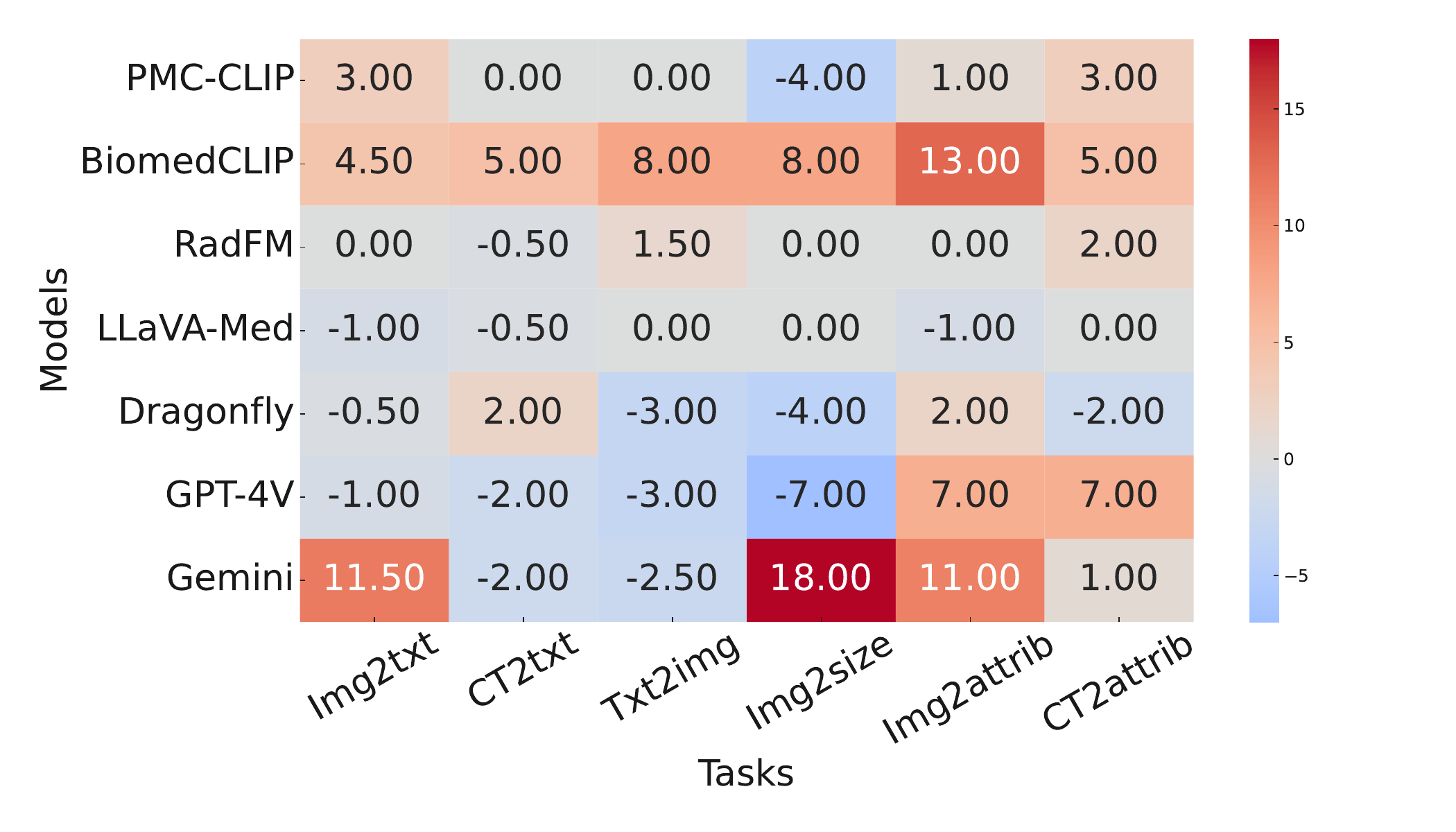}
    \caption{Heatmap of performance change with BBox: model performance differences across tasks when BBox annotations are used. Positive values indicate performance gains due to BBox, while negative values indicate performance drops.
}
    \label{improvement}
    \vspace{-10pt}  % 同样是可选
\end{figure}

Figure~\ref{improvement} summarizes the effect of BBox inputs across tasks and models. Providing BBox information consistently improves performance for Img2txt QA, Context2txt QA, Img2attrib QA, and CT2attrib QA, with particularly large gains observed for BiomedCLIP and Gemini. In contrast, Txt2img QA shows minimal sensitivity to BBox inputs, indicating that image retrieval primarily relies on global visual--text alignment rather than localized lesion cues. The impact of BBox on Img2size QA is mixed, with modest improvements for some models and negligible changes for others.

\begin{figure}[t]
    \vspace{-6pt}
    \centering
    \includegraphics[width=0.9\columnwidth]{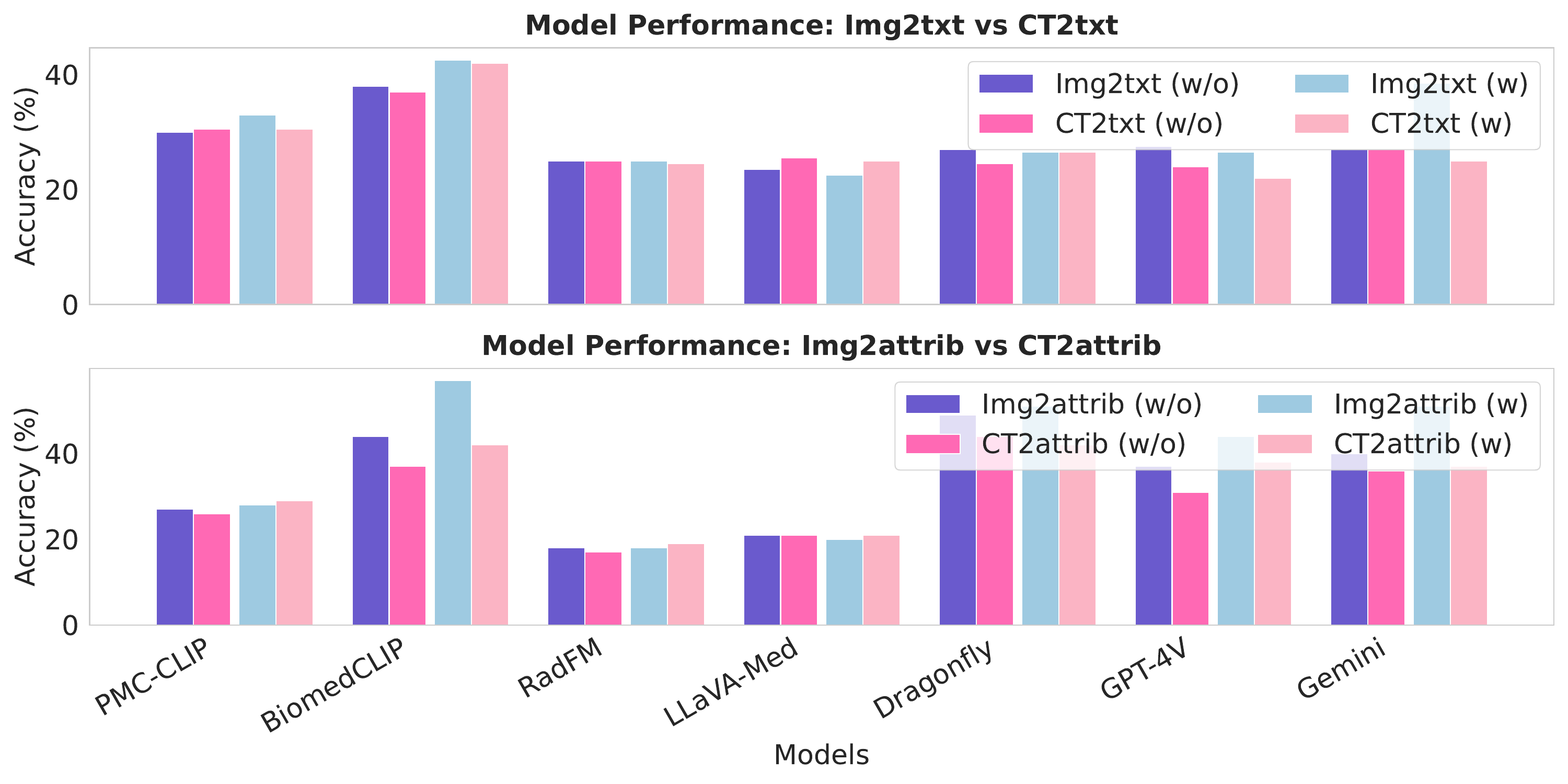}
    \caption{Comparison of model performance on single-slice (Img2txt / Img2attrib) versus multi-slice context tasks (Context2txt / Context2attrib).
}
    \label{ctvsimg}
    \vspace{-6pt}
\end{figure}

Figure~\ref{ctvsimg} compares performance between single-slice image tasks and multi-slice context-based tasks. Across all models, accuracy on standard single-slice tasks (e.g., Img2txt) is consistently higher than on context-based tasks (e.g., Context2txt). While models such as Gemini and BiomedCLIP perform well on single-slice inputs, their performance degrades substantially when multi-slice contextual reasoning is required, highlighting the increased difficulty of volumetric CT understanding.

\subsection{Human Evaluation Results}

To assess the clinical value of CT-Bench, we conducted a human evaluation with two senior radiologists and one junior doctor. Each reviewed 100 randomly chosen cases per task and condition. For Img2txt, Context2txt, and Txt2img, there were 10 examples each (with and without BBox). Txt2attrib, CT2attrib, and Img2size had 5 examples each per condition, while Txt2bbox was tested only with boxes (10 examples).
Results (Table \ref{Human Evaluation Results}) show that CT-Bench closely aligns with senior radiologists, especially on tasks involving BBox—reaching over 90\% agreement. The junior doctor had lower agreement, particularly on tasks without spatial cues. This highlights CT-Bench’s ability to reflect expert-level understanding and its sensitivity to clinical skill. All evaluators performed worse without BBox, underscoring the importance of spatial grounding in complex cases. Overall, CT-Bench proves to be a clinically meaningful, expert-aligned, and challenging benchmark for evaluating medical AI.

\section{Discussion}

The differential impact of bounding box supervision across tasks highlights important distinctions in how multimodal models utilize spatial information. Tasks that require explicit grounding between visual regions and language, such as Img2txt, Context2txt, and attribute classification, benefit most from bounding box inputs because spatial cues reduce ambiguity and constrain attention to lesion-relevant regions. In contrast, tasks such as Txt2img rely primarily on global semantic alignment between image and text embeddings, which may explain their limited sensitivity to localized supervision. The mixed effect observed for lesion size estimation further suggests that accurate geometric reasoning remains a challenge for current vision--language models, even when lesion boundaries are approximately localized.

The pronounced performance gap between single-slice and multi-slice tasks underscores the intrinsic difficulty of volumetric reasoning in CT imaging. Unlike natural images, CT data require integration of information across adjacent slices to capture lesion extent, morphology, and anatomical context. Most evaluated models process slices independently or rely on shallow aggregation strategies, which likely limits their ability to form coherent three-dimensional representations. These findings suggest that future multimodal systems may benefit from architectures explicitly designed for volumetric encoding, cross-slice attention mechanisms, or hybrid representations that better capture 3D spatial structure.

Furthermore, the observation that specialized models such as BiomedCLIP and Gemini derive greater benefit from bounding box supervision than generalist or lower-capacity models indicates that architectural readiness plays a critical role in exploiting spatial guidance. Models equipped with stronger attention mechanisms and domain-adapted pretraining appear better positioned to leverage localized supervision signals, whereas weaker models may be unable to translate such cues into meaningful performance gains. This highlights the importance of jointly considering dataset design and model architecture when developing multimodal CT reasoning systems.

CT-Bench, comprising the Lesion Image \& Metadata Set and the QA Benchmark Component, provides a comprehensive resource for advancing multimodal AI in CT-based lesion analysis. It includes 20,335 lesions with radiologist-verified descriptions and bounding boxes, together with a seven-task QA benchmark featuring hard negatives and expert-reviewed answers. This enables rigorous training and evaluation of clinically meaningful AI systems across captioning, grounding, retrieval, and contextual reasoning tasks.

Despite these advancements, several challenges remain. Models still perform substantially less effectively on CT-based tasks than on standard vision--language benchmarks derived from natural images, indicating that existing vision--language models are not yet well adapted to the unique spatial and contextual demands of volumetric medical imaging. The structured, layered nature of CT scans---particularly in multi-slice contexts---introduces complexity that most general-purpose and even medical-domain models struggle to handle effectively.

In addition, producing high-quality bounding box annotations remains resource-intensive, requiring substantial expert labor. Beyond simply identifying lesions, the annotation process often involves translating complex and sometimes ambiguous clinical descriptions into structured visual and textual representations. This limits the scalability of CT-Bench and slows dataset expansion and iterative development cycles. Future work may explore semi-automated or AI-assisted annotation pipelines to reduce this burden while maintaining clinical reliability.

Finally, even after fine-tuning, the strongest-performing model (BiomedCLIP) achieves only 62\% average accuracy on QA tasks with bounding box support, which remains well below expert-level performance. Performance degrades further in settings without spatial cues, underscoring that current multimodal AI systems are not yet sufficiently robust for autonomous clinical deployment, among other safety issues \cite{yang2024unmasking,yang2025adversarial,jin2024hidden}. These gaps highlight the need for CT-specific modeling strategies, improved volumetric reasoning architectures, and more scalable annotation approaches, with CT-Bench serving as a foundation for advancing next-generation multimodal medical AI systems.

\section{Method}
\label{sec:method}

\subsection{CT-Bench: Lesion Image \& Metadata Set}
\textbf{Data sources.} Images were sourced from the DeepLesion dataset, which provides lesion-level BBoxes. Lesion descriptions and size information were extracted and standardized from PACS radiology reports.  

\textbf{Annotation challenges.} Three key challenges were addressed: : (1) Lesion Description Complexity: Many lesion descriptions referenced multiple lesions in a single sentence, requiring precise disambiguation of references (see examples in Appendix)
(2) Terminology Consistency: Variations in terms (e.g., ``foci'' vs. ``focal lesion'', ``enlarged'' vs. ``largest'') needed to be standardized to maintain clarity.
(3)Annotator Variability: Annotator performance differed based on expertise and training background. For example, annotators with medical degrees (MDs) achieved higher accuracy compared to those with non-medical PhDs.  

\textbf{Annotation pipeline.} To address these challenges, we designed a multi-step annotation pipeline that integrates GPT-4 pre-annotations with human-in-the-loop refinement. The pipeline consists of the following stages: (1) Stage 1: Initial Structuring and GPT-4 Fine-Tuning:
    Annotators first created preliminary annotations for 200 CT cases and then refined them through collaborative discussion. These improved annotations were used to fine-tune GPT-4. (2) Stage 2: Iterative Annotation and Continuous Fine-Tuning:
   The fine-tuned GPT-4 generated preliminary annotations for 100 new CT cases per round. Annotators refined these annotations through three iterative feedback cycles, progressively improving accuracy. After each round, the refined annotations were used to further fine-tune GPT-4. (3) Stage 3: Large-Scale Annotation and Verification:
   The fine-tuned  GPT-4 model was deployed to annotate the remaining training, test, and validation datasets. A dual-layer review process was implemented in this process, where two annotators reviewed the annotations, followed by verification from a medical expert. A comprehensive explanation of each stage, along with performance statistics, review procedures, fine-tuning details, and flowcharts, can be found in Appendix .
   
\textbf{Dataset composition.} CT-Bench comprises 20,335 lesions from 7,795 CT studies (3,793 patients). The dataset is split into training (15,380 lesions), validation (1,758), and test (3,197) sets (Table~\ref{tab:dataset_summary}). For each lesion, we release windowed PNG slices (soft-tissue window, C=50, W=400), with and without bounding-box overlays, plus structured metadata.

% \subsection{CT-Trigent dataset}
% \input{figs/figure2}
%\input{figs/lesion_size}

\label{sec:ctbench}

\label{CT-Bench Dataset}

% \input{figs/case}
% This section provides task-specific examples for CT-Bench in Figure \ref{case}.
% We introduce \textbf{CT-Bench}, a suite of seven tasks that evaluate visual-textual understanding in clinical imaging. Here we summarize each task briefly. Complete details, including dataset construction, methodology, and figures (\ref{random_alternative}, \ref{new_sample}, \ref{hard_negative}), can be found in Appendix~\ref{sec:appendix-tasks}.

% Here we describe the method used to select the challenging examples...
\subsection{CT-Bench: QA Benchmark Component}

After constructing the CT-Bench: Lesion Image \& Metadata Set, we further developed its multiple-choice QA benchmark component, designed to evaluate multimodal AI performance on lesion-level CT interpretation. These cases are shown in Figure \ref{fig:template}(d).
CT-Bench provides a comprehensive set of question-answer pairs that reflect clinically relevant tasks grounded in real lesion annotations. Here we summarize each task briefly: (1) \textbf{Img2txt:} Identify the most appropriate lesion description for a given image. To construct the this, we selected a subset of cases from the Lesion Image \& Metadata Set, using their original descriptions as correct answers. The initial setting of this task is randomly select  the other three answers from the remaining test set. However, these alternatives
were too simple, making it easy for the model to identify
the correct option. To address this, for each image, we use the BiomedCLIP \cite{zhang2023biomedclip} model to  retrieve 20 visually
similar cases. 

\begin{figure}[t]
    \centering
    \includegraphics[width=\columnwidth]{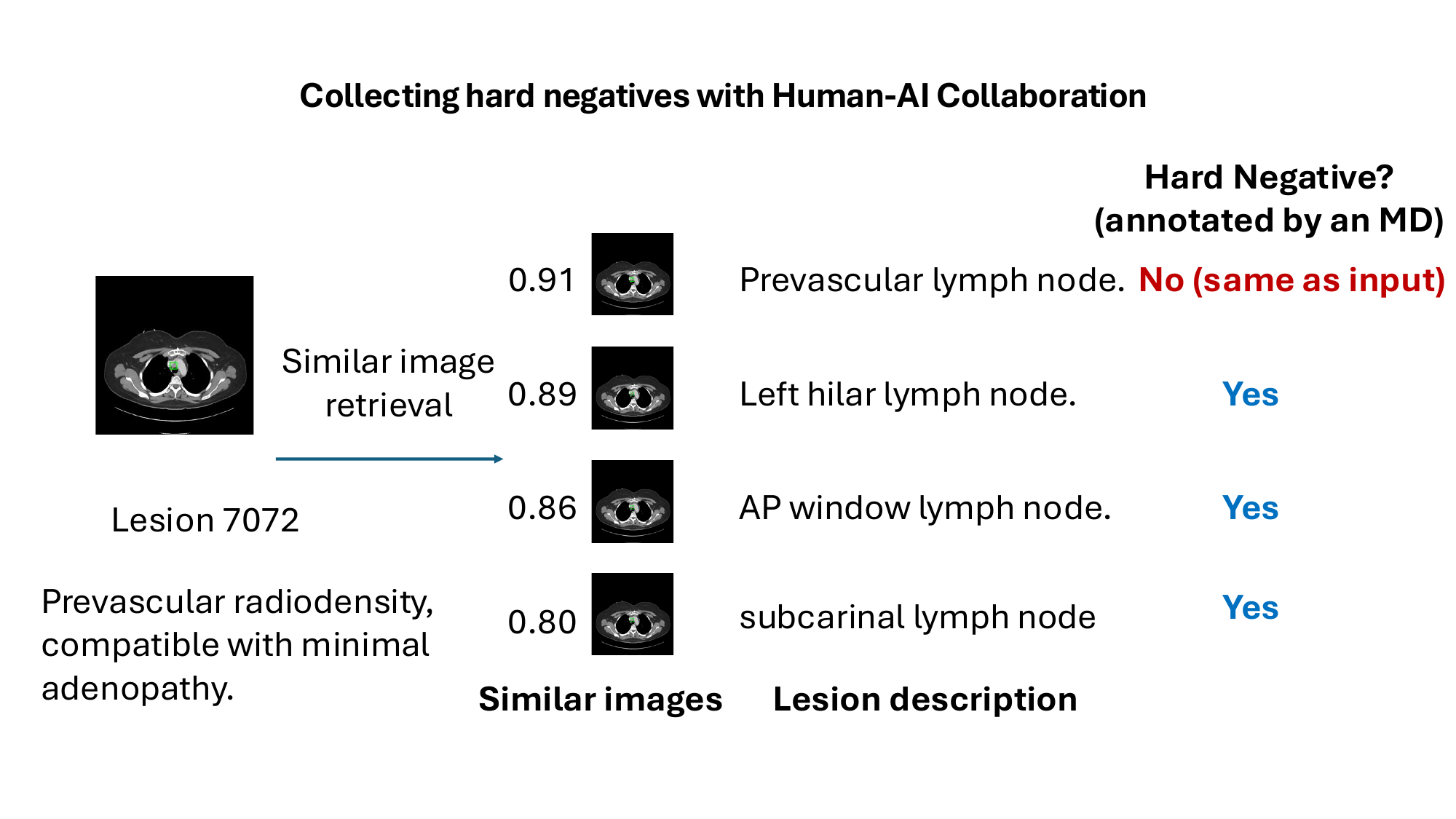}
    \caption{Illustration of the selection process for ``hard negative'' cases using the BiomedCLIP model and MD validation.}
    \label{hard_negative}
\end{figure}

From these, three ``hard negative'' cases were
selected by a medical doctor (MD). 
This process
is visualized in Figure \ref{hard_negative}. Examples of randomly selected and hard negative examples, along with the experimental setup, are provided in Appendix.
(2) \textbf{Context2txt:}  Identify the most appropriate lesion description for CT images. The question in this task is similar to Img2txt, but we concatenate nine consecutive CT slices (centered on the lesion), offering richer anatomical context to mimic radiological practice and improving the model’s lesion analysis. (3) \textbf{Txt2img:} Select the image that best matches a given lesion description.  We reuse the same \emph{hard negatives} from Img2txt to ensure visually challenging choices. (4) \textbf{Txt2bbox:} This task involves localizing lesions in images using BBoxes.  We sampled ground-truth BBoxes from the test dataset and used BiomedCLIP to retrieve visually similar images for each case. BBoxes were then detected in the retrieved images, with non-overlapping ones selected and inserted into the original images as hard negative examples. (5) \textbf{Img2size:} Estimate the size of a lesion (longer axis) in a CT image. The model must select a size category from predefined options, reflecting a key clinical assessment where lesion dimension often correlates with diagnostic importance.
(6) \textbf{Img2attrib:} Select lesion attributes based on an ontology (e.g., location, lesion type, and additional properties). First, we fine-tuned GPT-4 to map textual lesion descriptions to ontology terms (details in Appendix). Next, we applied BiomedCLIP to retrieve visually similar images and get their attributes. Finally, a medical doctor curated four attribute choices based on these results. (7) \textbf{Context2attrib:}  Select lesion attributes for CT images. This follows the same ontology-driven method as Img2attrib, but adapted for multi-slice CT data as described in Context2txt.
Overall, CT-Bench’s diversity in task type enables a broad evaluation of medical AI performance. 
% The QA benchmark is divided into validation and test sets, further categorized by the presence or absence of BBox annotations.
Table \ref{tab:dataset_distribution} shows the distribution of QA pairs.
For compact visualization in figures and tables, the multi-slice context tasks
\texttt{Context2txt} and \texttt{Context2attrib} are abbreviated as
\texttt{CT2txt} and \texttt{CT2attrib}, respectively.

% \begin{table*}[t]\small
% \centering
% \small
% \resizebox{\textwidth}{!}{%
% \begin{tabular}{lcccccccccccccccc}
% \toprule
% \textbf{Modality} & 
% \multicolumn{2}{c}{\textbf{Img2txt}} & 
% \multicolumn{2}{c}{\textbf{CT2txt}} & 
% \multicolumn{2}{c}{\textbf{Txt2img}} & 
% \multicolumn{2}{c}{\textbf{Txt2bbox}} & 
% \multicolumn{2}{c}{\textbf{Img2attrib}} & 
% \multicolumn{2}{c}{\textbf{CT2attrib}} & 
% \multicolumn{2}{c}{\textbf{Img2size}} & 
% \multicolumn{2}{c}{\textbf{Total}} \\ 
% \cmidrule(lr){2-3} \cmidrule(lr){4-5} \cmidrule(lr){6-7} \cmidrule(lr){8-9} 
% \cmidrule(lr){10-11} \cmidrule(lr){12-13} \cmidrule(lr){14-15} \cmidrule(lr){16-17}
% & \textbf{w/o} & \textbf{w/} 
% & \textbf{w/o} & \textbf{w/} 
% & \textbf{w/o} & \textbf{w/} 
% & \textbf{w/o} & \textbf{w/} 
% & \textbf{w/o} & \textbf{w/} 
% & \textbf{w/o} & \textbf{w/} 
% & \textbf{w/o} & \textbf{w/} 
% & \textbf{w/o} & \textbf{w/} \\
% \midrule
% \textbf{Validation} & 100 & 100 & 100 & 100 & 100 & 100 & - & 50 & 50 & 50 & 50 & 50 & 50 & 50 & 450 & 500 \\
% \textbf{Test} & 200 & 200 & 200 & 200 & 200 & 200 & - & 100 & 100 & 100 & 100 & 100 & 100 & 100 & 900 & 1,000 \\
% \bottomrule
% \end{tabular}
% }
% \caption{Distribution of the \textbf{CT-Bench: QA Benchmark Component} across seven VQA task types, split by validation and test sets, and by presence of BBox annotations. Abbreviations used:
% w/o = Without BBox, w/ = With BBox.}
% \label{tab:dataset_distribution}

% \end{table*}
\textbf{Hard negatives.} To increase task difficulty and simulate real-world ambiguity, we curated three types of distractors:  
(i) \textit{Appearance-based textual distractors}, retrieved via BiomedCLIP and MD-verified;  
(ii) \textit{Localization distractors}, generated from visually similar images and non-overlapping BBoxes;  
(iii) \textit{Ontology-driven attribute distractors}, constructed from RadLex mappings with expert curation.  

\subsection{Experimental Setup}

\noindent\textbf{Baseline Models} To advance multimodal AI in radiology, a diverse range of vision-language models were evaluated, spanning three major categories. (1) General Vision-Language Models such as GPT-4 \cite{achiam2023gpt} and Gemini \cite{team2023gemini} demonstrate broad image-text reasoning capabilities across domains. (2) Medical Vision-Language Models like LLaVA-Med \cite{li2024llava}, RadFM \cite{wu2023towards}, 
% BimedGPT \cite{zhang2024generalist}, 
and Dragonfly \cite{thapa2024dragonfly} are tailored to the medical domain, leveraging specialized datasets to enhance performance on radiology tasks. (3) Medical CLIP Models, including BiomedCLIP \cite{zhang2023biomedclip} and PMC-CLIP \cite{lin2023pmc}, utilize contrastive learning to align radiological images with medical text in a shared embedding space, improving retrieval and lesion identification. Input adaptations were made to accommodate each model's architecture, as detailed in Appendix.

\textbf{Model Training.}  
While these models have significantly advanced medical multimodal AI, the lack of a robust CT-based benchmark with fine-grained lesion descriptions has limited their applicability for real-world radiology AI development \cite{zhu2024decomposing}. To address this, we fine-tuned two representative models—\textbf{RadFM} and \textbf{BiomedCLIP}—from the medical VLM and CLIP categories, respectively, using the CT-Bench: Lesion Image \& Metadata Set. Both versions of the dataset—with and without Bbox annotations—were used in training. All fine-tuning was performed using PyTorch on a cluster of 32 NVIDIA A100 GPUs (80 GB memory each).
\textbf{RadFM Fine-Tuning:}  
RadFM was trained using a batch size of 4 and a 4-step gradient accumulation strategy. The maximum token length was set to 2048, and the model was fine-tuned for 4 epochs.
\textbf{BiomedCLIP Fine-Tuning:}  
For BiomedCLIP, we used the Adam optimizer with a learning rate of \(1 \times 10^{-5}\). The model was trained for 200 epochs using a batch size of 64. Cross-Entropy Loss was applied to address the multi-class classification nature of QA tasks , where the goal is to select the most appropriate textual description for a given image.

\noindent\textbf{Evaluation metrics}

Model performance was evaluated using task-specific metrics tailored to each component of CT-Bench. For the image captioning task in the Lesion Image \& Metadata Set, we used BLEU \cite{papineni2002bleu}, METEOR \cite{banerjee2005meteor}, ROUGE\cite{lin2004rouge}, BERTScore \cite{zhang2019bertscore} and Cosine Similarity \cite{zhang2019bertscore} based on Sentence-BERT embeddings \cite{reimers2019sentence}. 

For the QA Benchmark Component, we used accuracy as the primary evaluation metric, quantifying the proportion of correctly answered questions. This metric offers a straightforward and interpretable assessment of model performance across the seven lesion analysis tasks \cite{powers2011evaluation}.

\subsection{Ethical Statement}

The data utilized in this study were fully anonymized and compliant with the Health Insurance Portability and Accountability Act (HIPAA). Use of the data was approved by the Institutional Review Board (IRB), with a waiver of informed consent. Access to GPT-4V and GPT-4, online large language models, was conducted via Microsoft Azure services to ensure secure and privacy-compliant data handling.

\section*{Data availability}

The dataset used in this study is publicly available on Kaggle at:
\url{https://kaggle.com/datasets/cd1661d6d6aeab08b8eb99b58885b4489d76fc5ac07d5aac76cae577e6426e2f}.
The dataset is distributed under the Creative Commons Attribution-NonCommercial-ShareAlike 4.0 International License (CC BY-NC-SA 4.0).

\section*{Acknowledgements}
This research was supported by the Intramural Research Program of the National Library of Medicine and Clinical Center at the NIH.

\section*{Author Contributions}

\textbf{Conceptualization:} Q.Z., Q.J. \\
\textbf{Methodology:} Q.Z., Q.J., T.S.M. \\
\textbf{Data Curation:} Q.Z., Q.J., T.S.M., Z.W., Y.Y., M.S.-G., B.H. \\
\textbf{Annotation:} Q.Z., Q.J., Z.W., Y.Y., M.S.-G., B.H. \\
\textbf{Model Development and Experiments:} Q.Z., Q.J., B.H. \\
\textbf{Validation:} Q.Z., P.T.S.B., K.C.W., R.M.S. \\
\textbf{Writing -- Original Draft:} Q.Z., Q.J., B.H. \\
\textbf{Writing -- Review \& Editing:} Y.F., Q.J., B.H., R.G., T.S.M., K.C.W., R.M.S., Z.L. \\
\textbf{Supervision:} R.M.S., Z.L.

All authors approved the final manuscript.

\section*{Competing interests}
The authors declare no competing interests.
\newpage
\clearpage
\section*{Tables}

\begin{table*}[h]
    \small
    \centering
    \renewcommand{\arraystretch}{1.2}
    \setlength{\tabcolsep}{8pt}
        \caption{Comparison of medical imaging datasets.}
    \resizebox{\textwidth}{!}{%
    \begin{tabular}{@{}>
    {\raggedright\arraybackslash}p{0.20\textwidth}
                    >{\raggedright\arraybackslash}p{0.35\textwidth}
                    >{\raggedright\arraybackslash}p{0.25\textwidth}
                    >{\raggedright\arraybackslash}p{0.05\textwidth}
                    >{\raggedright\arraybackslash}p{0.25\textwidth}@{}}
        \toprule
        \textbf{Dataset} & \textbf{Image Data Type} & \textbf{Textual Annotations} & \textbf{Size} & \textbf{Notes} \\
        \midrule
        ROCOv2 & CT, MRI, PET, Ultrasound, X-ray & General captions & $\sim$80k & Image-text pairs \\
        PMC-OA & Biomedical images & Captions from PubMed & 1M+ & Image-text pairs \\
        CT-RATE & CT (Thoracic, non-contrast) & Basic radiology reports & 25k & 3D CT-report pairs \\
        DeepLesion & CT (BBox available) & No textual annotations & $\sim$32k & Images only \\
        MEDICAT & Radiology, histology, scope procedures & Captions, references & 7.5k &  subfigure-subcaption pairs \\
        \textbf{Ours \newline(Lesion\& Metadata) } & CT (Bbox available) & Lesion descriptions and size info from PACS& $\sim$20k & Image-text pairs \\
        \bottomrule
    \end{tabular}
    }

    \label{tab:medical_datasets}
\end{table*}
\begin{table*}[h]
    \small
    \centering
    \renewcommand{\arraystretch}{1.2} % More vertical padding
     \caption{Comparison of medical VQA datasets in terms of size, imaging modality, answer type, QA curation approach, and the inclusion of hard negatives. \textsuperscript{*}H.N. = Hard Negatives; Open = open-ended questions; MC = multiple choice.}
    \resizebox{\textwidth}{!}{%
    \begin{tabular}{l lllll}
        \toprule
        \textbf{Dataset} & \textbf{Size} \newline (Images) & \textbf{Modality} & \textbf{Answer Type} & \textbf{QA Curation} & \textbf{H.N.}\textsuperscript{*} \\
        \midrule
        VQA-RAD         & 0.3k   & MRI, CT, X-ray, etc.       & Open/MC   & Manually curated                                     & No  \\
        PathVQA         & 5k     & Microscopy Images          & Open      & Textbook-based, manually verified                   & No  \\
        SLAKE           & 0.7k   & MRI, CT, X-ray, etc.       & Open/MC   & Manually curated                                     & No  \\
        VQA-Med         & 5k     & MRI, CT, X-ray, etc.       & Open      & Manually curated                                     & No  \\
        PMC-VQA         & 149k   & Mixed                      & Open/MC   & ChatGPT-generated                                   & Yes \\
        \textbf{Ours (QA)} & 2.8k & CT (BBox available)       & MC        & Template-based, manually verified                   & Yes \\
        \bottomrule
    \end{tabular}
    }
   
    \label{tab:qa_datasets}
\end{table*}

\begin{table}\small
\caption{Summary of the CT-Bench: Lesion Image \& Metadata Set.}
\label{tab:dataset_summary}
\centering
\small
\setlength{\tabcolsep}{4pt}
\begin{tabular}{lrrrrr}
\toprule
\textbf{Dataset} & \textbf{Patients} & \textbf{Male} & \textbf{Female} & \textbf{Studies} & \textbf{Lesions} \\
\midrule
Training    & 2,835 & 1,516 & 1,319 & 5,875 & 15,380 \\
Validation  & 342   & 181   & 161   & 667   & 1,758  \\
Test        & 616   & 367   & 249   & 1,253 & 3,197  \\
\midrule
\textbf{Total} & \textbf{3,793} & \textbf{2,064} & \textbf{1,729} & \textbf{7,795} & \textbf{20,335} \\
\bottomrule
\end{tabular}
\end{table}

\begin{table*}[t]\small
\caption{Distribution of the \textbf{CT-Bench: QA Benchmark Component} across seven VQA task types, split by validation and test sets, and by presence of BBox annotations. Abbreviations used:
w/o = Without BBox, w/ = With BBox.}
\centering
\small
\resizebox{\textwidth}{!}{%
\begin{tabular}{lcccccccccccccccc}
\toprule
\textbf{Modality} & 
\multicolumn{2}{c}{\textbf{Img2txt}} & 
\multicolumn{2}{c}{\textbf{CT2txt}} & 
\multicolumn{2}{c}{\textbf{Txt2img}} & 
\multicolumn{2}{c}{\textbf{Txt2bbox}} & 
\multicolumn{2}{c}{\textbf{Img2attrib}} & 
\multicolumn{2}{c}{\textbf{CT2attrib}} & 
\multicolumn{2}{c}{\textbf{Img2size}} & 
\multicolumn{2}{c}{\textbf{Total}} \\ 
\cmidrule(lr){2-3} \cmidrule(lr){4-5} \cmidrule(lr){6-7} \cmidrule(lr){8-9} 
\cmidrule(lr){10-11} \cmidrule(lr){12-13} \cmidrule(lr){14-15} \cmidrule(lr){16-17}
& \textbf{w/o} & \textbf{w/} 
& \textbf{w/o} & \textbf{w/} 
& \textbf{w/o} & \textbf{w/} 
& \textbf{w/o} & \textbf{w/} 
& \textbf{w/o} & \textbf{w/} 
& \textbf{w/o} & \textbf{w/} 
& \textbf{w/o} & \textbf{w/} 
& \textbf{w/o} & \textbf{w/} \\
\midrule
\textbf{Validation} & 100 & 100 & 100 & 100 & 100 & 100 & - & 50 & 50 & 50 & 50 & 50 & 50 & 50 & 450 & 500 \\
\textbf{Test} & 200 & 200 & 200 & 200 & 200 & 200 & - & 100 & 100 & 100 & 100 & 100 & 100 & 100 & 900 & 1,000 \\
\bottomrule
\end{tabular}
}

\label{tab:dataset_distribution}

\end{table*}

\begin{table*}[t]\small
\footnotesize
\caption{Summary of performance for models with and without BBoxes. The highest values for unsupervised models are highlighted in bold, and the best fine-tuned models are underlined. }
\centering
{
\resizebox{\textwidth}{!}{%
\begin{tabular}{lcccccccccccccccccc}
\toprule
\textbf{Model} & 
\multicolumn{2}{c}{\textbf{BLEU-1}} & 
\multicolumn{2}{c}{\textbf{METEOR}} & 
\multicolumn{2}{c}{\textbf{ROUGE-1}} & 
\multicolumn{2}{c}{\textbf{BERT(F1)}} & \multicolumn{2}{c}{\textbf{CosineSim}}\\
\cmidrule(lr){2-3} \cmidrule(lr){4-5} \cmidrule(lr){6-7} \cmidrule(lr){8-9} \cmidrule(lr){10-11}
& \textbf{w/o} & \textbf{w/} 
& \textbf{w/o} & \textbf{w/} 
& \textbf{w/o} & \textbf{w/} 
& \textbf{w/o} & \textbf{w/} & \textbf{w/o} & \textbf{w/}\\
\midrule
Unsupervised\\

RadFM & 0.0713 & 0.0779 & 0.0984 & 0.1034 & 0.0573 & 0.0604& \textbf{0.8518} & 0.8525  &0.3262&0.3325\\
Dragonfly & \textbf{0.0842} & \textbf{0.1685} & 0.1045 & \textbf{0.2242} & 0.0307& \textbf{0.1250} &0.8430 &\textbf{0.8681}  & 0.2638&0.4082\\
Llava\_med & 0.0756& 0.0802 & 0.1555 & 0.1666 & 0.0748 &0.0804  & 0.8462 & 0.8484 &0.3914& 0.4014\\
GPT-4V & 0.0656 & 0.0916 & 0.1405 & 0.1915 & 0.0586 & 0.0819  & 0.8424 &0.8549 &0.3895 &0.4456 \\
Gemini & 0.0754 & 0.1021 & \textbf{0.1666} &0.2027  &\textbf{0.0709} & 0.0954 &0.8499 & 0.8582 &\textbf{0.4149}& \textbf{0.4730} \\
\hline
Fine-tuned\\

RadFM(w/o) & 0.2258 & - & 0.2142 & - & 0.1302 & - & 0.8811 & -& 0.4567& -\\
RadFM(w) &- &\underline{0.2448} & -& \underline{0.2382}& - & \underline{0.1507} & - & \underline{0.8878} &- & \underline{0.4911}  \\

\bottomrule
\end{tabular}}
}

\vspace{-0.05in}
\label{performance_comparison}
\end{table*}
% \noindent\textbf{Results on CT-Bench  QA Benchmark for all models} 
% Tables can be inserted via the normal table and tabular environment. To put
% footnotes inside tables you should use \verb+\footnotetext[]{...}+ tag.
% The footnote appears just below the table itself (refer Tables~\ref{tab1} and \ref{tab2}). 
% For the corresponding footnotemark use \verb+\footnotemark[...]+

% \begin{table}[h]
% \caption{Caption text}\label{tab1}%
% \begin{tabular}{@{}llll@{}}
% \toprule
% Column 1 & Column 2  & Column 3 & Column 4\\
% \midrule
% row 1    & data 1   & data 2  & data 3  \\
% row 2    & data 4   & data 5\footnotemark[1]  & data 6  \\
% row 3    & data 7   & data 8  & data 9\footnotemark[2]  \\
% \botrule
% \end{tabular}
% \footnotetext{Source: This is an example of table footnote. This is an example of table footnote.}
% \footnotetext[1]{Example for a first table footnote. This is an example of table footnote.}
% \footnotetext[2]{Example for a second table footnote. This is an example of table footnote.}
% \end{table}

\begin{table*}[t]
\caption{
Accuracy performance (\%) of different models across QA Benchmark tasks, reported with 95\% confidence intervals based on the binomial proportion for each task. 
Per-task results include the mean $\pm$ confidence interval. 
For aggregate metrics (\textbf{Average}), we report point estimates of the mean accuracy across tasks. 
Since the average spans heterogeneous tasks with different sample sizes, binomial confidence intervals are not directly applicable; instead, averages are presented as summary statistics for overall comparison. 
A Random baseline (25\%) is included as a reference for chance-level performance.
}
\centering
\small
\resizebox{\textwidth}{!}{%
\begin{tabular}{lccccccccccccccccc}
\toprule
\textbf{Model} & 
\multicolumn{2}{c}{\textbf{Img2txt}} & 
\multicolumn{2}{c}{\textbf{CT2txt}} & 
\multicolumn{2}{c}{\textbf{Txt2img}} & 
\multicolumn{2}{c}{\textbf{Txt2bbox}} & 
\multicolumn{2}{c}{\textbf{Img2attrib}} & 
\multicolumn{2}{c}{\textbf{CT2attrib}} & 
\multicolumn{2}{c}{\textbf{Img2size}} & 
\multicolumn{3}{c}{\textbf{Average}} \\ 
\cmidrule(lr){2-3} \cmidrule(lr){4-5} \cmidrule(lr){6-7} \cmidrule(lr){8-9} 
\cmidrule(lr){10-11} \cmidrule(lr){12-13} \cmidrule(lr){14-15} \cmidrule(lr){16-18}
& \textbf{w/o} & \textbf{w/} 
& \textbf{w/o} & \textbf{w/} 
& \textbf{w/o} & \textbf{w/} 
& \textbf{w/o} & \textbf{w/} 
& \textbf{w/o} & \textbf{w/} 
& \textbf{w/o} & \textbf{w/} 
& \textbf{w/o} & \textbf{w/} 
& \textbf{w/o} & \textbf{w/} & \textbf{Total} \\
\midrule
Unsupervised \\

Random & $25.0 \pm 6.0$ & $25.0 \pm 6.0$ & $25.0 \pm 6.0$ & $25.0 \pm 6.0$ & $25.0 \pm 6.0$ & $25.0 \pm 6.0$ & -- & $25.0 \pm 8.49$ & $25.0 \pm 8.49$ & $25.0 \pm 8.49$ & $25.0 \pm 8.49$ & $25.0 \pm 8.49$ & $25.0 \pm 8.49$ & $25.0 \pm 8.49$ & $25.0$ & $25.0$ & $25.0$ \\
PMC-CLIP & $30.0 \pm 6.35$ & $33.0 \pm 6.52$ & $30.5 \pm 6.38$ & $30.5 \pm 6.38$ & $24.5 \pm 5.96$ & $24.5 \pm 5.96$ & -- & $19.0 \pm 7.69$ & $27.0 \pm 8.7$ & $28.0 \pm 8.8$ & $26.0 \pm 8.6$ & $29.0 \pm 8.89$ & $20.0 \pm 7.84$ & $16.0 \pm 7.19$ & $27.0$ & $26.8$ & $26.9$ \\
BiomedCLIP & $\mathbf{38.0 \pm 6.73}$ & $\mathbf{42.5 \pm 6.85}$ & $\mathbf{37.0 \pm 6.69}$ & $\mathbf{42.0 \pm 6.84}$ & $26.5 \pm 6.12$ & $\mathbf{34.5 \pm 6.59}$ & -- & $35.0 \pm 9.35$ & $44.0 \pm 9.73$ & $\mathbf{57.0 \pm 9.7}$ & $37.0 \pm 9.46$ & $\mathbf{42.0 \pm 9.67}$ & $30.0 \pm 8.98$ & $38.0 \pm 9.51$ & $\mathbf{34.9}$ & $\mathbf{41.0}$ & $\mathbf{38.1}$ \\
RadFM & $25.0 \pm 6.0$ & $25.0 \pm 6.0$ & $25.0 \pm 6.0$ & $24.5 \pm 5.96$ & $3.5 \pm 2.55$ & $5.0 \pm 3.02$ & -- & $3.0 \pm 3.34$ & $18.0 \pm 7.53$ & $18.0 \pm 7.53$ & $17.0 \pm 7.36$ & $19.0 \pm 7.69$ & $0.0 \pm 0.0$ & $0.0 \pm 0.0$ & $15.8$ & $14.9$ & $15.3$ \\
LLaVA-Med & $23.5 \pm 5.88$ & $22.5 \pm 5.79$ & $25.5 \pm 6.04$ & $25.0 \pm 6.0$ & $0.0 \pm 0.0$ & $0.0 \pm 0.0$ & -- & $0.0 \pm 0.0$ & $21.0 \pm 7.98$ & $20.0 \pm 7.84$ & $21.0 \pm 7.98$ & $21.0 \pm 7.98$ & $0.0 \pm 0.0$ & $0.0 \pm 0.0$ & $15.6$ & $13.6$ & $14.5$ \\
Dragonfly & $27.0 \pm 6.15$ & $26.5 \pm 6.12$ & $24.5 \pm 5.96$ & $26.5 \pm 6.12$ & $29.5 \pm 6.32$ & $26.5 \pm 6.12$ & -- & $26.0 \pm 8.6$ & $\mathbf{49.0 \pm 9.8}$ & $51.0 \pm 9.8$ & $\mathbf{44.0 \pm 9.73}$ & $\mathbf{42.0 \pm 9.67}$ & $\mathbf{37.0 \pm 9.46}$ & $33.0 \pm 9.22$ & $32.4$ & $31.1$ & $31.7$ \\
GPT-4V & $27.5 \pm 6.19$ & $26.5 \pm 6.12$ & $24.0 \pm 5.92$ & $22.0 \pm 5.74$ & $28.0 \pm 6.22$ & $25.0 \pm 6.0$ & -- & $30.0 \pm 8.98$ & $37.0 \pm 9.46$ & $44.0 \pm 9.73$ & $31.0 \pm 9.06$ & $38.0 \pm 9.51$ & $30.0 \pm 8.98$ & $23.0 \pm 8.25$ & $28.6$ & $28.2$ & $28.4$ \\
Gemini & $27.0 \pm 6.15$ & $38.5 \pm 6.74$ & $27.0 \pm 6.15$ & $25.0 \pm 6.0$ & $\mathbf{32.5 \pm 6.49}$ & $30.0 \pm 6.35$ & -- & $\mathbf{36.0 \pm 9.41}$ & $40.0 \pm 9.6$ & $51.0 \pm 9.8$ & $36.0 \pm 9.41$ & $37.0 \pm 9.46$ & $32.0 \pm 9.14$ & $\mathbf{50.0 \pm 9.8}$ & $31.2$ & $36.1$ & $33.8$ \\
% Qwen & 21.5&25&26&26.5&29&33.5&&38&42&45&32&36&39&24&31.58&32.57&32\\
\hline
Fine-tuned \\

RadFM(w/o) & $0.0 \pm 0.0$ & $0.0 \pm 0.0$ & $0.0 \pm 0.0$ & $0.0 \pm 0.0$ & $0.0 \pm 0.0$ & $0.0 \pm 0.0$ & -- & $0.0 \pm 0.0$ & $0.0 \pm 0.0$ & $0.0 \pm 0.0$ & $0.0 \pm 0.0$ & $0.0 \pm 0.0$ & $0.0 \pm 0.0$ & $0.0 \pm 0.0$ & $0.0$ & $0.0$ & $0.0$ \\
RadFM(w) & $0.0 \pm 0.0$ & $0.0 \pm 0.0$ & $0.0 \pm 0.0$ & $0.0 \pm 0.0$ & $0.0 \pm 0.0$ & $0.0 \pm 0.0$ & -- & $0.0 \pm 0.0$ & $0.0 \pm 0.0$ & $0.0 \pm 0.0$ & $0.0 \pm 0.0$ & $0.0 \pm 0.0$ & $0.0 \pm 0.0$ & $0.0 \pm 0.0$ & $0.0$ & $0.0$ & $0.0$ \\
BiomedCLIP(w/o) & \underline{$42.0 \pm 6.84$} & $42.0 \pm 6.84$ & \underline{$41.0 \pm 6.82$} & $43.0 \pm 6.86$ & $32.0 \pm 6.47$ & $36.0 \pm 6.65$ & -- & $29.0 \pm 8.89$ & \underline{$54.0 \pm 9.77$} & $45.0 \pm 9.75$ & \underline{$46.0 \pm 9.77$} & $48.0 \pm 9.79$ & \underline{$43.0 \pm 9.7$} & $36.0 \pm 9.41$ & \underline{41.4} & $40.0$ & $40.7$ \\
BiomedCLIP(w) & $38.5 \pm 6.74$ & \underline{$68.0 \pm 6.47$} & $40.5 \pm 6.8$ & \underline{$62.0 \pm 6.73$} & $29.0 \pm 6.29$ & \underline{$57.5 \pm 6.85$} & -- & \underline{$67.0 \pm 9.22$} & $44.0 \pm 9.73$ & \underline{$67.0 \pm 9.22$} & $44.0 \pm 9.73$ & \underline{$65.0 \pm 9.35$} & $30.0 \pm 8.98$ & $46.0 \pm 9.77$ & $37.1$ & \underline{62.0} & \underline{50.3} \\
% Qwen(w/o)&34&33&32.5&36.5&26&26.5& &25& 48&56&47&46&39&46&35.44&36.50&36.00\\ Qwen(w)& 
% 26&48.50&27.50&47.00&28.50&43&&44&43&63&36& 63&22&90&29.44&53.70&42.21\\
% Qwen\_CT(w/o)& 
% 27&&&&29&&&&&53&47& 50&53&&&&\\

\bottomrule
\end{tabular}
}

\label{tab:model_performance_weighted}
\end{table*}

% \begin{table}[h]
% \centering
% \small
% \begin{tabular}{lcc}
% \toprule
% \textbf{Model} & \textbf{Avg. Inference Time (s)} & \textbf{Deployment Type} \\
% \midrule
% BiomedCLIP & $\approx$ 0.06 & Local (A100 GPU) \\
% PMC-CLIP & $\approx$ 0.06 & Local (A100 GPU) \\
% RadFM & $\approx$ 0.77 & Local (A100 GPU) \\
% LLaVA-Med & $\approx$ 2.14 & Local (A100 GPU) \\
% Dragonfly & $\approx$ 4.29 & Local (A100 GPU) \\
% GPT-4V & $\approx$ 2.98 & API (Azure Cloud) \\
% Gemini & $\approx$ 2.98 & API (Google Cloud) \\
% \bottomrule
% \end{tabular}
% \caption{Average inference time of baseline models. 
% GPT-4V and Gemini are accessed via cloud APIs, and their latency includes network and server queue time. 
% These figures are not directly comparable to local inference times.}
% \label{tab:inference_times}
% \end{table}

\begin{table*}[t]
\caption{
Human evaluation accuracy (\%) across CT-Bench tasks by clinical role. 
}
\centering
\small
\resizebox{\textwidth}{!}{%
\begin{tabular}{lccccccccccccccccc}
\toprule
\textbf{Role} & 
\multicolumn{2}{c}{\textbf{Img2txt}} & 
\multicolumn{2}{c}{\textbf{CT2txt}} & 
\multicolumn{2}{c}{\textbf{Txt2img}} & 
\multicolumn{2}{c}{\textbf{Txt2bbox}} & 
\multicolumn{2}{c}{\textbf{Img2attrib}} & 
\multicolumn{2}{c}{\textbf{CT2attrib}} & 
\multicolumn{2}{c}{\textbf{Img2size}} & 
\multicolumn{3}{c}{\textbf{Average}} \\ 
\cmidrule(lr){2-3} \cmidrule(lr){4-5} \cmidrule(lr){6-7} \cmidrule(lr){8-9}
\cmidrule(lr){10-11} \cmidrule(lr){12-13} \cmidrule(lr){14-15} \cmidrule(lr){16-18}
& \textbf{w/o} & \textbf{w/} 
& \textbf{w/o} & \textbf{w/} 
& \textbf{w/o} & \textbf{w/} 
& \textbf{w/o} & \textbf{w/} 
& \textbf{w/o} & \textbf{w/} 
& \textbf{w/o} & \textbf{w/} 
& \textbf{w/o} & \textbf{w/} 
& \textbf{w/o} & \textbf{w/} & \textbf{Total} \\
\midrule
Senior Radiologist 1 & 
100.0 & 90.0 & 
80.0 & 60.0 & 
80.0 & 70.0 & 
-- & 100.0 & 
60.0 & 100.0 & 
40.0 & 100.0 & 
40.0 & 100.0 & 
66.0 & 90.0 & 79.20 \\
Senior Radiologist 2 & 
70.0 & 100.0 & 
80.0 & 70.0 & 
80.0 & 70.0 & 
-- & 100.0 & 
60.0 & 100.0 & 
60.0 & 100.0 & 
80.0 & 100.0 & 
70.5 & 90.5 & 81.50 \\
Junior Doctor (Radiology Exp.) & 
50.0 & 80.0 & 
70.0 & 50.0 & 
30.0 & 40.0 & 
-- & 80.0 & 
60.0 & 80.0 & 
40.0 & 100.0 & 
0.0 & 40.0 & 
45.5 & 68.5 & 58.15 \\
\bottomrule
\end{tabular}
}

\label{Human Evaluation Results}
\end{table*}

\begin{table*}[!hbt]
\caption{
Comparison of model outputs for lesion localization and description.
}
\small
\centering
\resizebox{\textwidth}{!}{%

\begin{tabular}{p{3cm} p{5.5cm} p{1.5cm} p{7cm}}
\toprule
\textbf{Model} & \textbf{Output} & \textbf{Alignment} & \textbf{Key Observations} \\
\midrule
\textbf{Ground Truth} & A nodule in the superior segment of the left lower lobe. & - & - \\
\midrule
\textbf{RadFM} & There are multiple nodular opacities scattered throughout the bilateral lungs. The largest is in the right upper lobe with a draining vessel. & \textbf{Misaligned} & Overinterprets findings (multiple nodules, feeding vessels) and incorrectly identifies a larger nodule in the right upper lobe. \\
\midrule
\textbf{TunedRadFM(w)} & Left lower lobe nodule. & \textbf{Most aligned} & Correctly identifies a nodule in the left lower lobe but lacks the specific location (superior segment). \\
\midrule
\textbf{LLAVA-med} & The image is a computed tomography (CT) scan of the chest. It shows a large mass in the right upper lobe of the lung. & \textbf{Misaligned} & Mislocalizes the finding to the right upper lobe and describes it as a "large mass" rather than a nodule. \\
\midrule
\textbf{GPT} & The image shows a transverse section of a chest CT scan with a well-circumscribed, round, hyperdense lesion in the right lung parenchyma. & \textbf{Misaligned} & Incorrectly identifies the lesion in the right lung; lacks diagnostic detail or specificity. \\
\midrule
\textbf{Gemini} & There is a small, well-defined nodule in the right lower lobe. The nodule is approximately 5 mm in diameter. No other significant findings. & \textbf{Misaligned} & Mislocalizes the nodule to the right lower lobe; adds size information not in the ground truth. \\
\midrule
\textbf{Dragonfly} & CT scan of the chest showing a 1.1 cm nodule in the left lower lobe. & \textbf{Most aligned} & Correctly identifies a nodule in the left lower lobe, but does not specify the superior segment; includes size information (1.1 cm), which is not in the ground truth. \\
\bottomrule
\end{tabular}

}

\vspace{-0.05in}
\label{tab:model_comparison}
\end{table*}
\noindent

\clearpage
\newpage

\section{Figures}

\backmatter
\newpage
\clearpage
% \setcounter{page}{1}
% \bmhead{Supplementary information}
\bibliography{sn-article}% common bib file

@article{zhu2024decomposing,
  title={How well do multimodal LLMs interpret CT scans? An auto-evaluation framework for analyses},
  author={Zhu, Qingqing and Hou, Benjamin and Mathai, Tejas Sudarshan and Mukherjee, Pritam and Jin, Qiao and Chen, Xiuying and Wang, Zhizheng and Cheng, Ruida and Summers, Ronald M and Lu, Zhiyong},
  journal={Journal of Biomedical Informatics},
  pages={104864},
  year={2025},
  publisher={Elsevier}
}

@article{thapa2024dragonfly,
  title={Dragonfly: Multi-Resolution Zoom-In Encoding Enhances Vision-Language Models},
  author={Thapa, Rahul and Chen, Kezhen and Covert, Ian and Chalamala, Rahul and Athiwaratkun, Ben and Song, Shuaiwen Leon and Zou, James},
  journal={arXiv preprint arXiv:2406.00977},
  year={2024}
}

@article{sun2023scoping,
  title={A scoping review on multimodal deep learning in biomedical images and texts},
  author={Sun, Zhaoyi and Lin, Mingquan and Zhu, Qingqing and Xie, Qianqian and Wang, Fei and Lu, Zhiyong and Peng, Yifan},
  journal={Journal of Biomedical Informatics},
  pages={104482},
  year={2023},
  publisher={Elsevier}
}

@article{goodfellow2013empirical,
  title={An empirical investigation of catastrophic forgetting in gradient-based neural networks},
  author={Goodfellow, Ian J and Mirza, Mehdi and Xiao, Da and Courville, Aaron and Bengio, Yoshua},
  journal={arXiv preprint arXiv:1312.6211},
  year={2013}
}

@inproceedings{papineni2002bleu,
  title={Bleu: a method for automatic evaluation of machine translation},
  author={Papineni, Kishore and Roukos, Salim and Ward, Todd and Zhu, Wei-Jing},
  booktitle={Proceedings of the 40th annual meeting of the Association for Computational Linguistics},
  pages={311--318},
  year={2002}
}

@inproceedings{banerjee2005meteor,
  title={METEOR: An automatic metric for MT evaluation with improved correlation with human judgments},
  author={Banerjee, Satanjeev and Lavie, Alon},
  booktitle={Proceedings of the acl workshop on intrinsic and extrinsic evaluation measures for machine translation and/or summarization},
  pages={65--72},
  year={2005}
}

@article{reimers2019sentence,
  title={Sentence-bert: Sentence embeddings using siamese bert-networks},
  author={Reimers, Nils and Gurevych, Iryna},
  journal={arXiv preprint arXiv:1908.10084},
  year={2019}
}

@inproceedings{lin2004rouge,
  title={Rouge: A package for automatic evaluation of summaries},
  author={Lin, Chin-Yew},
  booktitle={Text summarization branches out},
  pages={74--81},
  year={2004}
}

@article{ruckert2024rocov2,
  title={Rocov2: Radiology objects in context version 2, an updated multimodal image dataset},
  author={R{\"u}ckert, Johannes and Bloch, Louise and Br{\"u}ngel, Raphael and Idrissi-Yaghir, Ahmad and Sch{\"a}fer, Henning and Schmidt, Cynthia S and Koitka, Sven and Pelka, Obioma and Abacha, Asma Ben and G. Seco de Herrera, Alba and others},
  journal={Scientific Data},
  volume={11},
  number={1},
  pages={688},
  year={2024},
  publisher={Nature Publishing Group UK London}
}

@article{hamamci2024foundation,
  title={A foundation model utilizing chest ct volumes and radiology reports for supervised-level zero-shot detection of abnormalities},
  author={Hamamci, Ibrahim Ethem and Er, Sezgin and Almas, Furkan and Simsek, Ayse Gulnihan and Esirgun, Sevval Nil and Dogan, Irem and Dasdelen, Muhammed Furkan and Wittmann, Bastian and Simsar, Enis and Simsar, Mehmet and others},
  journal={CoRR},
  year={2024}
}

@article{achiam2023gpt,
  title={Gpt-4 technical report},
  author={Achiam, Josh and Adler, Steven and Agarwal, Sandhini and Ahmad, Lama and Akkaya, Ilge and Aleman, Florencia Leoni and Almeida, Diogo and Altenschmidt, Janko and Altman, Sam and Anadkat, Shyamal and others},
  journal={arXiv preprint arXiv:2303.08774},
  year={2023}
}

@article{team2023gemini,
  title={Gemini: a family of highly capable multimodal models},
  author={Team, Gemini and Anil, Rohan and Borgeaud, Sebastian and Alayrac, Jean-Baptiste and Yu, Jiahui and Soricut, Radu and Schalkwyk, Johan and Dai, Andrew M and Hauth, Anja and Millican, Katie and others},
  journal={arXiv preprint arXiv:2312.11805},
  year={2023}
}

@article{li2024llava,
  title={Llava-med: Training a large language-and-vision assistant for biomedicine in one day},
  author={Li, Chunyuan and Wong, Cliff and Zhang, Sheng and Usuyama, Naoto and Liu, Haotian and Yang, Jianwei and Naumann, Tristan and Poon, Hoifung and Gao, Jianfeng},
  journal={Advances in Neural Information Processing Systems},
  volume={36},
  year={2024}
}

@article{wu2023towards,
  title={Towards generalist foundation model for radiology},
  author={Wu, Chaoyi and Zhang, Xiaoman and Zhang, Ya and Wang, Yanfeng and Xie, Weidi},
  journal={arXiv preprint arXiv:2308.02463},
  year={2023}
}

@article{zhang2023biomedclip,
  title={BiomedCLIP: a multimodal biomedical foundation model pretrained from fifteen million scientific image-text pairs},
  author={Zhang, Sheng and Xu, Yanbo and Usuyama, Naoto and Xu, Hanwen and Bagga, Jaspreet and Tinn, Robert and Preston, Sam and Rao, Rajesh and Wei, Mu and Valluri, Naveen and others},
  journal={arXiv preprint arXiv:2303.00915},
  year={2023}
}

@article{tian2024opportunities,
  title={Opportunities and challenges for ChatGPT and large language models in biomedicine and health},
  author={Tian, Shubo and Jin, Qiao and Yeganova, Lana and Lai, Po-Ting and Zhu, Qingqing and Chen, Xiuying and Yang, Yifan and Chen, Qingyu and Kim, Won and Comeau, Donald C and others},
  journal={Briefings in Bioinformatics},
  volume={25},
  number={1},
  pages={bbad493},
  year={2024},
  publisher={Oxford University Press}
}

@article{johnson2019mimic,
  title={MIMIC-CXR-JPG, a large publicly available database of labeled chest radiographs},
  author={Johnson, Alistair EW and Pollard, Tom J and Greenbaum, Nathaniel R and Lungren, Matthew P and Deng, Chih-ying and Peng, Yifan and Lu, Zhiyong and Mark, Roger G and Berkowitz, Seth J and Horng, Steven},
  journal={arXiv preprint arXiv:1901.07042},
  year={2019}
}

@inproceedings{lin2023pmc,
  title={Pmc-clip: Contrastive language-image pre-training using biomedical documents},
  author={Lin, Weixiong and Zhao, Ziheng and Zhang, Xiaoman and Wu, Chaoyi and Zhang, Ya and Wang, Yanfeng and Xie, Weidi},
  booktitle={International Conference on Medical Image Computing and Computer-Assisted Intervention},
  pages={525--536},
  year={2023},
  organization={Springer}
}

@article{subramanian2020medicat,
  title={Medicat: A dataset of medical images, captions, and textual references},
  author={Subramanian, Sanjay and Wang, Lucy Lu and Mehta, Sachin and Bogin, Ben and van Zuylen, Madeleine and Parasa, Sravanthi and Singh, Sameer and Gardner, Matt and Hajishirzi, Hannaneh},
  journal={arXiv preprint arXiv:2010.06000},
  year={2020}
}

@article{lau2018dataset,
  title={A dataset of clinically generated visual questions and answers about radiology images},
  author={Lau, Jason J and Gayen, Soumya and Ben Abacha, Asma and Demner-Fushman, Dina},
  journal={Scientific data},
  volume={5},
  number={1},
  pages={1--10},
  year={2018},
  publisher={Nature Publishing Group}
}

@article{he2020pathvqa,
  title={Pathvqa: 30000+ questions for medical visual question answering},
  author={He, Xuehai and Zhang, Yichen and Mou, Luntian and Xing, Eric and Xie, Pengtao},
  journal={arXiv preprint arXiv:2003.10286},
  year={2020}
}

@inproceedings{liu2021slake,
  title={Slake: A semantically-labeled knowledge-enhanced dataset for medical visual question answering},
  author={Liu, Bo and Zhan, Li-Ming and Xu, Li and Ma, Lin and Yang, Yan and Wu, Xiao-Ming},
  booktitle={2021 IEEE 18th International Symposium on Biomedical Imaging (ISBI)},
  pages={1650--1654},
  year={2021},
  organization={IEEE}
}

@inproceedings{ben2021overview,
  title={Overview of the vqa-med task at imageclef 2021: Visual question answering and generation in the medical domain},
  author={Ben Abacha, Asma and Sarrouti, Mourad and Demner-Fushman, Dina and Hasan, Sadid A and M{\"u}ller, Henning},
  booktitle={Proceedings of the CLEF 2021 Conference and Labs of the Evaluation Forum-working notes},
  year={2021},
  organization={21-24 September 2021}
}

@article{zhang2023pmc,
  title={Pmc-vqa: Visual instruction tuning for medical visual question answering},
  author={Zhang, Xiaoman and Wu, Chaoyi and Zhao, Ziheng and Lin, Weixiong and Zhang, Ya and Wang, Yanfeng and Xie, Weidi},
  journal={arXiv preprint arXiv:2305.10415},
  year={2023}
}

@article{powers2011evaluation,
  title={Evaluation: From precision, recall and F-measure to ROC, informedness, markedness and correlation},
  author={Powers, David MW},
  journal={Journal of Machine Learning Technologies},
  volume={2},
  number={1},
  pages={37--63},
  year={2011}
}

@article{zhang2019bertscore,
  title={BERTScore: Evaluating text generation with BERT},
  author={Zhang, Tianyi and Kishore, Varsha and Wu, Felix and others},
  journal={arXiv preprint arXiv:1904.09675},
  year={2019}
}

@article{hou2025one,
  title={One year on: assessing progress of multimodal large language model performance on RSNA 2024 case of the day questions},
  author={Hou, Benjamin and Mukherjee, Pritam and Batheja, Vivek and Wang, Kenneth C and Summers, Ronald M and Lu, Zhiyong},
  journal={Radiology},
  volume={316},
  number={2},
  pages={e250617},
  year={2025},
  publisher={Radiological Society of North America}
}

@article{jin2024hidden,
  title={Hidden flaws behind expert-level accuracy of multimodal GPT-4 vision in medicine},
  author={Jin, Qiao and Chen, Fangyuan and Zhou, Yiliang and Xu, Ziyang and Cheung, Justin M and Chen, Robert and Summers, Ronald M and Rousseau, Justin F and Ni, Peiyun and Landsman, Marc J and others},
  journal={NPJ Digital Medicine},
  volume={7},
  number={1},
  pages={190},
  year={2024},
  publisher={Nature Publishing Group UK London}
}

@article{yang2025adversarial,
  title={Adversarial prompt and fine-tuning attacks threaten medical large language models},
  author={Yang, Yifan and Jin, Qiao and Huang, Furong and Lu, Zhiyong},
  journal={Nature Communications},
  volume={16},
  number={1},
  pages={9011},
  year={2025},
  publisher={Nature Publishing Group UK London}
}

@article{yang2024unmasking,
  title={Unmasking and quantifying racial bias of large language models in medical report generation},
  author={Yang, Yifan and Liu, Xiaoyu and Jin, Qiao and Huang, Furong and Lu, Zhiyong},
  journal={Communications medicine},
  volume={4},
  number={1},
  pages={176},
  year={2024},
  publisher={Nature Publishing Group UK London}
}

@article{zhu2025well,
  title={How well do multimodal LLMs interpret CT scans? An auto-evaluation framework for analyses},
  author={Zhu, Qingqing and Hou, Benjamin and Mathai, Tejas Sudarshan and Mukherjee, Pritam and Jin, Qiao and Chen, Xiuying and Wang, Zhizheng and Cheng, Ruida and Summers, Ronald M and Lu, Zhiyong},
  journal={Journal of Biomedical Informatics},
  pages={104864},
  year={2025},
  publisher={Elsevier}
}

@article{yan2017deeplesion,
  title={Deeplesion: Automated deep mining, categorization and detection of significant radiology image findings using large-scale clinical lesion annotations},
  author={Yan, Ke and Wang, Xiaosong and Lu, Le and Summers, Ronald M},
  journal={arXiv preprint arXiv:1710.01766},
  year={2017}
}
\newpage
\clearpage
\setcounter{page}{1}

\end{document}